\documentclass{article}

\usepackage[final,nonatbib]{neurips_2026}
\makeatletter
\renewcommand{\@noticestring}{The 10th Machine Learning for Systems Workshop at NeurIPS 2026, Sydney, Australia.}
\makeatother

\usepackage[utf8]{inputenc} 
\usepackage[T1]{fontenc}    
\usepackage{hyperref}       
\usepackage{url}            
\usepackage{booktabs}       
\usepackage{amsfonts}       
\usepackage{amsmath}        
\usepackage{nicefrac}       
\usepackage{microtype}      
\usepackage{xcolor}         
\usepackage{graphicx}       
\usepackage{multirow}       
\usepackage{array}          

\title{Intelligent Embedding Pruning for Large-Scale Recommendation Models}

\author{%
Anish Khazane\thanks{Author emails, in author order: \texttt{\{akhazane, alexsaadfalcon, guanchu, mflporto, jaspershan, feiy, eugeneshulga, intaik, yongzhe, xiaoyud, luos, xiaoya213\}@meta.com}.} \\ Meta
\And Alex Saad-Falcon \\ Meta
\And Guanchu Wu \\ Meta
\And Marcio Porto \\ Meta
\AND Jasper Shan \\ Meta
\And Fei Yu \\ Meta
\And Eugene Shulga \\ Meta
\And Intaik Park \\ Meta
\AND Yongzhe Zhang \\ Meta
\And Xiaoyu Deng \\ Meta
\And Sheng Luo \\ Meta
\And Xiaoya Wang \\ Meta
}

\begin{document}

\maketitle
\begin{abstract}

Modern recommendation systems face substantial memory pressure, largely from the sparse embedding layer that maps high-cardinality categorical features into dense vectors. In-training embedding pruning (ITEP) removes less important embedding rows during training, yielding end-to-end memory savings but sometimes degrading accuracy because per-table pruning ratios are set by size-based heuristics. We use feature importance (FI) to guide these ratios (FI-ITEP), assigning higher pruning to tables tied to less important features. Across three large-scale recommendation models, FI-ITEP achieves 20--40\% total memory reduction while matching or improving accuracy relative to percentile-based ITEP. Because the feature-importance signal requires minimal resources to compute, these memory savings translate into concrete systems benefits: the same models train on fewer GPU trainers (e.g., 112 to 80) with higher per-GPU throughput, saving 1,000+ GPU-hours per model launch. We reproduce these trends on the open-source DLRM (Criteo 1TB) and quantify the end-to-end training-time and GPU-cost tradeoff---including the burden of computing FI---offering practical guidance for embedding pruning in real-world systems.
\end{abstract}

\section{Introduction}

Recommendation systems power personalized experiences across e-commerce, social media, and content streaming with models spanning billions of parameters. In distributed training, GPU high-bandwidth memory (HBM)---not compute---is often the binding constraint, so larger models drive up GPU count, resource use, and energy. The embedding layer, which maps high-cardinality categorical features (user IDs, post IDs, engineered features) into dense vectors, dominates this footprint.

Embedding pruning shrinks this footprint by removing less important or infrequently used rows. In-training embedding pruning (ITEP), introduced as AdaEmbed \cite{adaembed}, extends pruning into the training loop, cutting resource sink for frequently retrained models. Its main limitation is heavy tuning: each sparse table can have its own pruning ratio, and the common percentile heuristic---prune larger tables more---ignores each feature's contribution to quality. As \cite{adaembed} notes, using feature-level information is a promising direction. Our contribution is FI-ITEP: using feature importance to set per-table pruning ratios, assigning higher ratios to less important features. By freeing additional HBM beyond the original ITEP, FI-ITEP lets us train the same models on fewer GPU trainers---e.g., from 112 to 80 GPUs with 8\% higher per-GPU throughput---saving 1,000+ GPU-hours per model launch, well beyond the 4--11\% overhead of computing FI (Section~\ref{sec:cost}).

\section{Method}

\subsection{Feature Importance}
\label{sec:methods_fi}

We use two perturbation-based FI techniques. \textbf{Permutation (shuffle)} \cite{breiman_random_2001} randomly shuffles a feature's values within a batch and measures the resulting change in model loss; larger changes indicate higher importance. \textbf{Shapley Value Sampling (SVS)} \cite{castro_polynomial_2009} approximates Shapley values by sampling feature permutations and averaging each feature's marginal contribution. We assign higher ITEP pruning ratios to embedding tables with lower FI scores, and denote this strategy FI-ITEP. The appendix (Fig.~\ref{fig:fi}) shows example FI distributions computed with shuffle and SVS for an industry-scale model.

\subsection{FI-ITEP Mechanism}

Following \cite{adaembed}, ITEP shrinks embedding tables by a target pruning ratio and maintains a lookup table mapping original IDs to pruned rows plus a frequency table tracking access patterns; less important IDs are hashed to a shared row as training progresses. The layer is lightweight, modular, and supports distributed training. In FI-ITEP, we use precomputed FI scores (Section~\ref{sec:methods_fi}) to set per-table pruning ratios, achieving 20--40\% total memory reduction while keeping train and eval NE on par with or better than size-based ITEP. Table~\ref{tab:model_summary} summarizes the three industry-scale models and the improvement of FI-ITEP over percentile-based ITEP.

\begin{table*}[t]
\centering
\renewcommand{\arraystretch}{1.2}
\resizebox{\textwidth}{!}{%
\begin{tabular}{lccccc}
\toprule
\textbf{Model} & \multicolumn{1}{c}{\textbf{\# of Sparse Features}} &
\multicolumn{1}{c}{\textbf{Raw Emb Size}} &
\multicolumn{1}{c}{\textbf{\# of GPUs}} &
\multicolumn{2}{c}{\textbf{FI-ITEP (Shuffle)}} \\
\cmidrule(lr){5-6}
& (Approximate Value) & (Approximate Value) & &
\textbf{Memory Saving} & \textbf{Avg. Eval NE Gain (\%)} \\
\midrule
Rec. Model 1 & 555 & 390 GB & 32  & $\approx$ 13.6\% & 36.4\% \\
Rec. Model 2  & 1012  & 1.1 TB & 128  & $\approx$ 14.2\% & 51.3\% \\
Rec. Model 3  & 423 & 80 GB   & 32  & $\approx$ 8.3\% & 32.0\% \\
\bottomrule
\end{tabular}%
}
\caption{\textit{A summary of the impact of FI-ITEP on DLRMs of varying sizes. The "Memory Saving" and "Avg. Eval NE Gain (\%)" columns indicate relative percentage improvements over the percentile-based ITEP for these models.}}
\label{tab:model_summary}
\end{table*}

\subsection{Data, Models, and Metric}

Each DLRM comprises sparse embedding layers, feature-interaction layers, dense layers, and one or more task-specific output heads, implemented in PyTorch for distributed multi-GPU training. Accuracy is measured by Normalized Entropy (NE); lower is better, and a relative NE delta of 0.02\% is considered significant \cite{necomparison1}, \cite{necomparison2}. Experiments used 16--128 Nvidia H100 GPUs per model, with roughly 3--5 days of training and 12--16 hours of evaluation per experiment.

\section{FI-ITEP on Recommendation Models}

\begin{figure*}[ht]

\begin{center}
\includegraphics[width=0.31\textwidth]{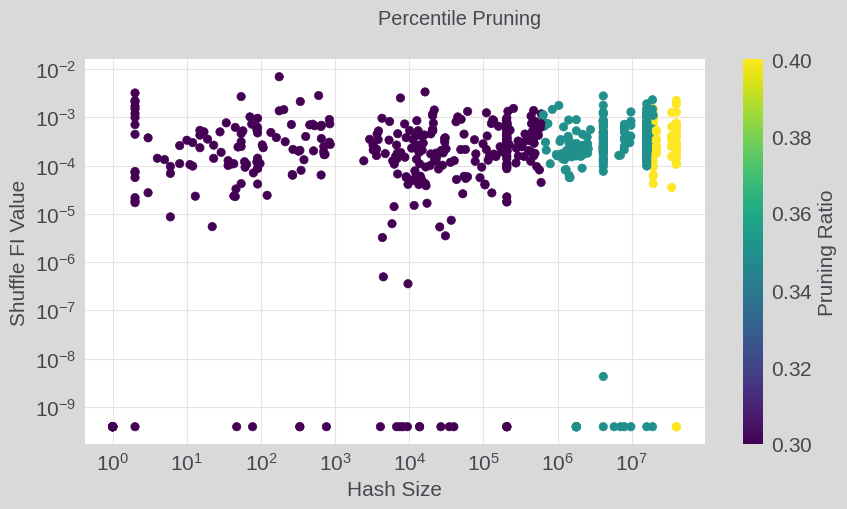}
\includegraphics[width=0.31\textwidth]{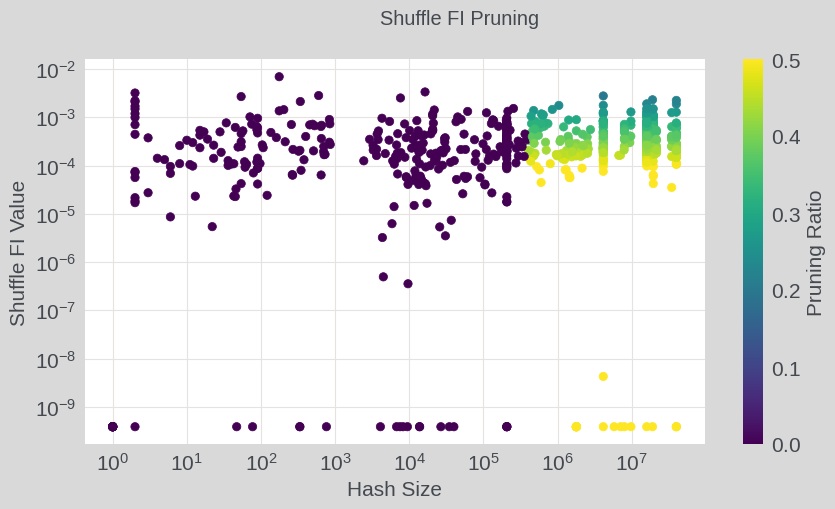}
\includegraphics[width=0.31\textwidth]{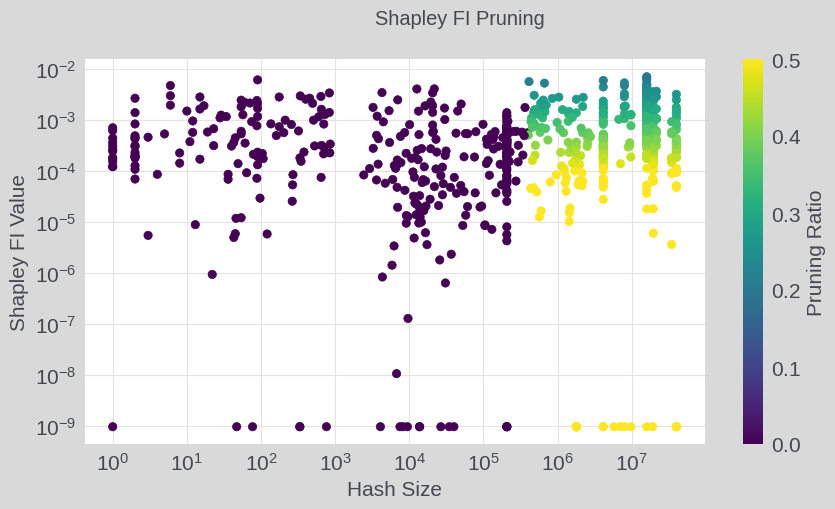}
\end{center}

\caption{\textit{Example pruning schedules for Recommendation Model 1 under the percentile, shuffle-FI, and SVS-FI strategies; color encodes each table's assigned pruning ratio. Embedding tables with low feature importance above a predefined minimum table size are assigned higher pruning ratios.}}
\label{fig:pruning_schedule}

\end{figure*}

We compare a percentile baseline against FI-ITEP on three industry-scale DLRMs of varying size (Table~\ref{tab:model_summary}). The percentile baseline (a wrapper around \cite{adaembed}) sorts embedding tables by size and assigns higher pruning ratios to larger tables---pruning the smallest tables lightly and the largest most aggressively---independent of each feature's importance. For each model we sweep pruning aggressiveness (10--50\% sparse, 5--40\% total) and run 3--5 repetitions near 50\% sparse reduction, reporting the train/eval NE gap versus an unpruned baseline (negative is better). Fig.~\ref{fig:pruning_schedule} shows example schedules under the percentile, shuffle-FI, and SVS-FI strategies: the percentile schedule hardcodes pruning ratios to discrete groups of table sizes (a banded, stepwise pattern), whereas shuffle-FI and SVS-FI linearly interpolate each table's ratio from its feature-importance score (a smooth gradient).

\textbf{Model 1 (medium, 390 GB).} At high aggressiveness, FI-ITEP reaches about 28.4\% total memory reduction (vs. 25\% for percentile) with an eval NE regression of about 1.4\%, versus about 2.2\% for percentile. In relative terms, that is a 13.6\% improvement in memory reduction and a 36.4\% smaller eval NE regression than percentile (Table~\ref{tab:model_summary}). Shapley (SVS) FI was evaluated only on Model 1, where Shuffle FI outperformed it with lower variance; tuning SVS is more delicate because it samples an exponential space of feature coalitions, so Models 2 and 3 use Shuffle FI only.

\textbf{Model 2 (large, over 1 TB).} FI-ITEP reaches about 32\% total memory reduction (vs. 28\% percentile). Both strategies improve NE, but FI-ITEP is the clear winner in evaluation, with an eval NE gain of -0.059\% versus -0.039\% for percentile. This is a 14.2\% relative memory-reduction gain and a 51.3\% larger eval NE improvement over percentile.

\textbf{Model 3 (small, under 100 GB).} FI-ITEP reaches about 34.98\% total memory reduction (vs. 32.3\% percentile). Both strategies show some NE regression on this smaller model, but FI-ITEP regresses less (eval NE about 0.017\% vs. 0.025\%). In relative terms, that is an 8.3\% memory-reduction gain and a 32.0\% smaller NE regression than percentile.

\textbf{Overall.} FI-ITEP performs best on the larger Models 1 and 2, with both greater memory savings and better eval NE than percentile. Larger models learn many low-contribution embeddings; FI identifies the important ones, enabling precise pruning that also acts as regularization. On the smaller Model 3 the benefit persists but is less pronounced, with fewer redundant embeddings to prune.

\section{Reproducibility on Open-Source Data}

To validate FI-ITEP, we evaluate it on the unmodified open-source DLRM \cite{dlrm} using the public Criteo 1TB dataset \cite{criteo} (Day 3, obtained from HuggingFace), comparing against percentile ITEP at matched memory-reduction levels. FI-ITEP consistently matches or outperforms percentile ITEP (Table~\ref{tab:criteo_auroc}), with the largest AUROC gains at the 50\% reduction level and a lower training budget---reproducing on a standard benchmark the trends observed on industry scale models. At identical memory footprints, FI-ITEP also delivers competitive or better training-time speedups (Table~\ref{tab:criteo_speedup}): at 80\% reduction it trains 29\% faster than baseline versus 14\% for percentile ITEP at the same footprint.

\begin{table}[h]
\centering
\begin{tabular}{ccccc}
\toprule
\textbf{Memory Reduction} & \textbf{Batches} & \textbf{FI-ITEP AUROC} & \textbf{Percentile-ITEP AUROC} & \textbf{$\Delta$ AUROC} \\
\midrule
50\% & 200  & 0.6395 & 0.6176 & +0.0218 \\
50\% & 1000 & 0.6920 & 0.6851 & +0.0068 \\
80\% & 200  & 0.6268 & 0.6223 & +0.0045 \\
80\% & 1000 & 0.6932 & 0.6932 & +0.0001 \\
\bottomrule
\end{tabular}
\vspace{2mm}
\caption{\textit{FI-ITEP vs. Percentile-ITEP on the open-source DLRM (Criteo 1TB) at matched memory reduction. ``Memory Reduction'' is the fraction of sparse-embedding memory removed.}}
\label{tab:criteo_auroc}
\end{table}

\begin{table}[h]
\centering
\begin{tabular}{lccc}
\toprule
\textbf{Configuration} & \textbf{Memory} & \textbf{Training Time} & \textbf{Speedup vs. Baseline} \\
\midrule
Baseline      & 1{,}476 MB & 130.7s & --- \\
FI-ITEP 50\%  & 738 MB     & 107.1s & 18\% faster \\
Percentile-ITEP 50\%  & 738 MB     & 107.7s & 18\% faster \\
FI-ITEP 80\%  & 295 MB     & 93.4s  & 29\% faster \\
Percentile-ITEP 80\%  & 295 MB     & 112.2s & 14\% faster \\
\bottomrule
\end{tabular}
\vspace{2mm}
\caption{\textit{Training-time speedup on the open-source DLRM (1000 batches) at matched memory footprint.}}
\label{tab:criteo_speedup}
\end{table}

\section{End-to-End Time and Computational Cost Tradeoff}\label{sec:cost}

Beyond raw memory reduction, we quantify the end-to-end training computational cost impact on a frequently retrained model. Lowering per-GPU HBM lets us run with fewer trainers that would otherwise hit out-of-memory failures. Reducing the allocation from 112 to 80 trainers lowers global P90 throughput (985K to 760K QPS) but increases per-GPU efficiency (8,800 to 9,501 P90 QPS/GPU, +8\%) due to larger local batches (Table~\ref{tab:gpu_savings}), saving roughly 108 GPU-hours per 12-hour-plus run. Because this model is retrained frequently for A/B testing and research cycles, aggregate savings reach 1,000+ GPU-hours per model launch, far outweighing the cost of computing FI.

That FI computational cost is modest (Table~\ref{tab:fi_overhead}): FI is computed on a fixed sample of 4,096 examples per batch over 256 batches per day of training data (only 1--2 GB regardless of model size), adding roughly 4--11\% to training time. FI scores are also stable across retraining---over a one-month interval the top-100 features have Spearman correlation 0.85--0.92 with under 12\% turnover (Table~\ref{tab:fi_drift}; Appendix D)---and since large recommenders are redeployed more frequently than monthly, drift is unlikely to be problematic in practice. As a rule of thumb, FI-ITEP is most worthwhile when projected fleet-wide resource savings exceed the roughly constant FI computation cost; since trainer-reduction savings shrink for smaller models while FI computational cost stays flat, standard size-based pruning may suffice when savings fall below cost.

\begin{table}[h]
\centering
\resizebox{\textwidth}{!}{%
\begin{tabular}{lccc}
\toprule
\textbf{Metric} & \textbf{Baseline (Unpruned)} & \textbf{FI-ITEP (96 Trainers)} & \textbf{FI-ITEP (80 Trainers)} \\
\midrule
Number of trainers (GPUs)   & 112  & 96   & 80   \\
P90 Global QPS              & 985K & 879K & 760K \\
P90 QPS/GPU                & 8{,}800 & 9{,}161 & 9{,}501 \\
P90 Per-GPU QPS \%          & 0\%  & 4\%  & 8\%  \\
Equivalent GPU-Hour Saving  & 0 hr & 54 hr & 108 hr \\
GPU Max Utilization         & 75\% & 80\% & 93\% \\
\bottomrule
\end{tabular}%
}
\vspace{2mm}
\caption{\textit{End-to-end GPU savings with FI-ITEP via trainer reduction on an industry-scale model.}}
\label{tab:gpu_savings}
\end{table}

\begin{table}[h]
\centering
\begin{tabular}{lccc}
\toprule
\textbf{Metric} & \textbf{Model 1} & \textbf{Model 2} & \textbf{Model 3} \\
\midrule
Total Sparse Features & 555 & 1012 & 423 \\
Total training data   & 55 GB & 80 GB & 50 GB \\
FI data sampled       & 1--2 GB & 1--2 GB & 1--2 GB \\
FI computation time   & 1h 37m & 1h 19m & 45m \\
Full run duration     & 38h & 11h 58m & 15h 20m \\
Total FI overhead     & 4.9\% & 11\% & 4.3\% \\
Num GPUs (H100)       & 4 & 14 & 2 \\
\bottomrule
\end{tabular}
\vspace{2mm}
\caption{\textit{Feature Importance (FI) computation overhead across industry-scale models.}}
\label{tab:fi_overhead}
\end{table}

\section{Conclusion and Future Work}

Feature importance is an effective, data-driven signal for setting per-table ITEP pruning ratios: FI-ITEP beats the percentile baseline on train and eval NE across model sizes while delivering 20--40\% total memory reduction. By cutting memory further than size-based heuristics such as percentile pruning, FI-ITEP shrinks the HBM footprint of large recommendation models enough to train them on fewer GPUs, directly lowering the hardware burden and energy of industry-scale RecSys training. These gains reproduce on the open-source Criteo benchmark and yield favorable end-to-end GPU-cost tradeoffs net of FI overhead. Future work includes better SVS tuning, longer-horizon study of the regularization effect, and extension to other architectures and hardware.

\appendix

\section*{Appendix}

The following supplementary material provides the full detail, figures, and per-model breakdowns condensed in the main text.

\section{Related Work}
\subsection{Feature Importance}

Feature importance (FI) refers to a broad set of algorithms used for computing the effect of different model inputs on making an accurate prediction. FI algorithms can either be data-driven or statistical, with data-driven algorithms offering more useful results for deep learning models.
Feature importance rankings are a foundational tool in machine learning, supporting a range of practical applications:

\begin{itemize}
\item
\textbf{Feature Selection:} By quantifying the relevance of individual features, practitioners can systematically remove those that do not substantially contribute to model performance \cite{theng_feature_2024}. This process reduces both training and inference computational costs while preserving or enhancing predictive accuracy.
\item
\textbf{Feature Deprecation:} Building on feature selection, features identified as globally unimportant can be removed from the system, resulting in further resource and capacity savings \cite{yu2004efficient}.
\item
\textbf{Model Explainability:} Feature importance helps interpret model outputs by identifying the most influential features for individual predictions. This approach enables practitioners to understand model decisions, diagnose unexpected behaviors, and build trust in the system's outputs \cite{anand_explainable_2022}.
\item
\textbf{Monitoring and Debugging:} In online or recurring training scenarios, feature importance metrics can be monitored over time to detect anomalies such as performance regressions, feature drift, or corruption, enabling timely intervention to mitigate issues \cite{siirtola2023featureRelevance}.
\end{itemize}

A diverse set of methodologies exists for estimating feature importance, each presenting unique advantages and limitations. Perturbation-based techniques evaluate the effect of systematically altering feature values on model performance. Features whose perturbation yields negligible changes in predictions are considered low-importance, whereas those causing substantial performance degradation are deemed highly important. This procedure is repeated across all features (or feature groups) to construct a comprehensive importance ranking. While perturbation-based techniques are model-agnostic and straightforward to implement, their computational demands scale poorly with the number of features, posing challenges for large-scale applications. Alternatively, gradient-based techniques such as Saliency Maps \cite{simonyan_deep_2014} and Integrated Gradients \cite{sundararajan_axiomatic_2017} leverage model gradients to estimate the sensitivity of predictions to input features. These techniques are often more computationally efficient, but may require access to model internals and can be less robust across heterogeneous model architectures.

\subsection{Embedding Pruning}

Post-training pruning (PTP) is a common method to reduce model memory demand during inference by reducing the size of sparse embedding tables after training. This technique has been applied to fully connected and convolutional neural networks \cite{han_learning_2015} and transformers \cite{kwon_fast_2022}.

In-training embedding pruning (ITEP) was discussed in the prior work under the name AdaEmbed \cite{adaembed}.
Deep learning recommendation models (DLRMs) rely on massive embedding tables to represent categorical features, but the ever-growing size of these tables leads to high deployment costs and slower execution \cite{hugectr}, \cite{xie2022fleche}. AdaEmbed introduces ITEP, an algorithm designed to reduce the size of embeddings required for DLRMs without sacrificing accuracy. ITEP leverages the observation that both the access patterns and the importance of different embeddings are highly heterogeneous and dynamic during training. By monitoring training time access frequencies and training gradients, ITEP identifies and retains only the most important embeddings, dynamically pruning less significant ones as training progresses. This approach allows more efficient embedding learning that can automatically adapt to training data distribution over time, reducing manual tuning for optimal resource usage.

ITEP is implemented as a lightweight shim layer that can be integrated into existing DLRM systems with minimal code changes. It features a distributed architecture with a central coordinator and on-worker agents that track embedding importance and enforce pruning decisions. The system introduces a Virtually Hashed Physically Indexed (VHPI) design to efficiently manage embedding memory and minimize overhead during pruning. Evaluations in industrial settings demonstrate that ITEP can reduce embedding size by 35-60\% and improve model execution speed by 11-34\%, all while maintaining or even improving model accuracy. By adaptively optimizing embeddings during training, ITEP not only saves resources but also achieves better generalization and reduces the need for manual configuration in large-scale recommendation systems. For more details, please refer to the original paper \cite{adaembed}.

The main shortcoming of ITEP is the extensive hyperparameter tuning required. Each sparse embedding table can have its own pruning ratio. An initial heuristic was to sort the tables by hash size and assign larger pruning ratios to larger tables. However, this does not account for the relative contributions of each sparse feature to model performance. The authors of \cite{adaembed} explicitly suggest incorporating feature-level information (e.g., feature importance) could be a promising direction, which we explore in this work.

\section{Extended Methods and Feature Importance Details}

\subsection{Feature Importance}
\label{sec:methods_fi}

For this work, we employ two perturbation-based feature importance techniques for improving the effectiveness of ITEP: shuffle (permutation) and Shapley values.
Permutation \cite{breiman_random_2001} works by randomly shuffling the values of a given feature within a batch and measuring the resulting change in the model output or loss (which serves as an indicator of the feature's importance). This process is analogous to feature ablation, where a feature is removed or replaced with a fixed value (such as zero) to assess its importance. Unlike traditional ablation, permutation ensures that each feature is replaced with a random value drawn from the original feature's data distribution to provide a more realistic assessment of feature importance.

Shapley Value Sampling (SVS) \cite{castro_polynomial_2009} is a practical approach for estimating Shapley values, which formally measure the contribution of each feature to a model's prediction. Exact computation of Shapley values \cite{kuhn_17_1953} is generally intractable for large-scale models due to the exponential number of feature combinations. To address this, Shapley Value Sampling constructs random permutations of the feature set and incrementally adds each feature to a baseline input. The change in model output observed after each addition represents the feature's marginal contribution in that permutation. By averaging these marginal contributions across multiple sampled permutations, the method yields an approximate attribution score for each feature, reflecting its overall influence on the model's output.

\begin{figure}[ht]

\begin{center}
\includegraphics[width=0.4\textwidth]{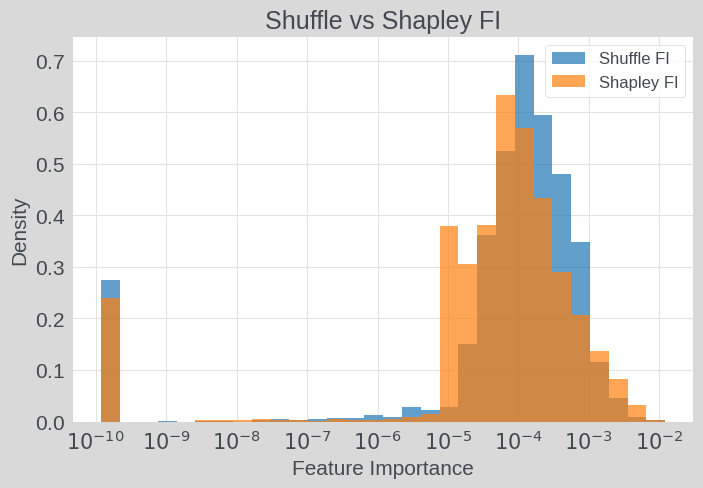}
\end{center}
\caption{\textit{Example feature importance values computed using shuffle and Shapley Value Sampling (SVS) for Recommendation Model 1. A floor is applied to raise all zero feature importance values to the minimum nonzero feature importance value.}}
\vspace{-1em}
\label{fig:fi}
\end{figure}

Fig. \ref{fig:fi} compares shuffle and SVS FI sampling on a large-scale recommendation model with thousands of sparse features. We see that SVS assigns slightly lower feature importance values for tables in comparison to Shuffle, likely due to capturing the impact of other signals like feature correlation. Accordingly, we can assign higher ITEP pruning ratios to embedding tables with lower feature importance scores, and denote this strategy in our experiments as FI-ITEP.

\subsection{Data and Model Architectures}
\label{sec:data_model_arch}

The accuracy of each DLRM model is measured by Normalized Entropy (NE) loss. A smaller NE indicates better performance, and any relative 0.02\% NE delta is considered to be significant \cite{necomparison1}, \cite{necomparison2}.

The models evaluated in this work are based on widely adopted deep learning architectures for large-scale recommendation systems. At a high level, each model comprises:
\begin{itemize}
\item
\textbf{Sparse Embedding Layers:} These layers map categorical features to dense vector representations, forming the primary source of memory consumption in the models.
\item
\textbf{Feature Interaction Layers:} Embedding outputs are combined using operations such as concatenation or element-wise product, followed by additional neural network layers to capture complex feature interactions.
\item
\textbf{Dense Neural Network Layers:} The combined feature representations are processed through fully connected layers, enabling the model to learn non-linear relationships and make accurate predictions.
\item
\textbf{Output Layer:} The final layer produces a prediction (or multiple predictions) on key tasks for the model in production.
\end{itemize}
All models were implemented using PyTorch and optimized for distributed training across multiple GPUs. We used three industry-grade recommendation models to test the efficacy of different pruning techniques on real-world models and data.

\subsection{In-Training Embedding Pruning}

ITEP was implemented on numerous models in a manner similar to \cite{adaembed}. The mechanism first shrinks the model's embedding tables by a specified pruning ratio prior to the start of training and creates a separate lookup table to map IDs from the previously unpruned table to the pruned one, along with a frequency table to track the access patterns for each of the original IDs. Then, IDs in the embedding table that are deemed less important—determined by the frequency table—are pruned by being hashed to the zeroth row of the embedding table. The lookup table update is performed iteratively, allowing the model to adaptively retain unique representations for only the most critical embeddings as training progresses. The pruning ratio and schedule are configurable, enabling experiments with different memory reduction targets and accuracy trade-offs. Furthermore, the ITEP layer is designed to be lightweight and modular, requiring minimal changes to the underlying model code and supporting distributed training across multiple GPUs.

In our approach, we use precomputed feature importance scores (FI-ITEP), as described in Section \ref{sec:methods_fi}, to define the pruning ratios per embedding table for each recommendation model. From integrating FI-ITEP into the training pipeline, we achieve further reductions in total memory usage --- between 20\% and 40\% --- while maintaining training and evaluation NE that's on-par with the original ITEP procedure proposed by \cite{adaembed}. This approach enables scalable deployment of DLRMs and provides a practical framework for balancing resource constraints with model performance.

\subsection{Hardware}

The experiments for this paper used numerous Nvidia H100 GPUs in parallel for each recommendation model. The number of GPUs per model ranged from 16 to 128. Each model was trained for approximately 3-5 days per experiment, with evaluation being done in 12-16 hours.

\section{Detailed Per-Model Results}

We will summarize the key empirical results of this work in the subsequent sections. We tested variations of in-training embedding pruning (ITEP) on three large-scale recommendation models of varying sizes: Recommendation Model 1 (medium, 390 GB), Recommendation Model 2 (large, 1.1TB), Recommendation Model 3 (small, 80 GB). A summary of these models is presented in Table \ref{tab:model_summary}.

For each model, we first tabulate the number of high-cardinality sparse features, which typically ranges from 50-1000. We apply a percentile pruning strategy as a baseline - a wrapper around \cite{adaembed} - where the smallest tables are pruned lightly, medium sized tables are pruned slightly more, and large tables are pruned the most. The percentile pruning strategy is an effective heuristic-based pruning strategy that has been observed to give good results out of the box.

For each model, we test an alternative pruning strategy: computing pruning ratios based on feature importance (FI) values described in Section~\ref{sec:methods_fi}. The pruning ratios for each sparse feature are computed automatically given two parameters: a minimum and maximum pruning ratio. The feature importance values are arranged on a logarithmic scale, with low feature importance values being assigned higher pruning ratios. As an example, if feature importance ranges from $10^{-5}$ to $10^{-1}$ with a minimum/maximum ratio set to 0.1 and 0.5, then sparse features with a feature importance of $10^{-2}$ will get assigned a pruning ratio of 0.2.

For Recommendation Model 1, we also experiment with two different feature importance algorithms: shuffle and Shapley value sampling (SVS). We observe better performance using shuffle, so for all subsequent models we only use shuffle FI values for ITEP configuration. We show example pruning ratio schedules for each of the different strategies in Fig. \ref{fig:pruning_schedule}.

For each recommendation model we perform the following experiments:
\begin{itemize}
\item
\textbf{Sweeping Pruning Aggressiveness:}

We sweep the aggressiveness of the ITEP pruning schedule to get a net sparse memory reduction ranging from 10\% to 50\%, leading to a total memory reduction ranging from 5\% to 40\% for the three tested models (a model's total HBM typically includes sparse, dense, and model parameters). For each aggressiveness level, we train the recommendation model on the same data and use two pruning schedules determined by percentile and feature importance. We report results for train and evaluation performance.
\item

\textbf{Repetitions on High Pruning Aggressiveness:}
We run 3-5 repetitions of model training on high pruning aggressiveness (close to 50\% sparse memory reduction) for each of the two pruning schedules in order to validate the impact on memory utilization and model quality. We report results for train and evaluation performance as well as variance across different training runs.

\end{itemize}

As discussed in Section~\ref{sec:data_model_arch}, we evaluate the impact of each version of ITEP on the normalized entropy (NE) of each DLRM. Lower NE is better, so a negative NE gap indicates the model has improved performance over the baseline. To reduce noise, the NE gap is averaged over numerous samples in training and evaluation and across different trained versions of the model.

\subsection{Recommendation Model 1}

Recommendation Model 1 is a medium DLRM that is about 390 GB in size with one primary recommendation task. We evaluate aggressive sparse embedding pruning (up to 50\% sparse HBM reduction) which moderately impacted model performance. On average, Percentile-ITEP yields 25\% HBM savings on this model, and FI-ITEP observed a total memory reduction of around 28.4\%. These results are summarized in Table \ref{tab:model_summary}.

\subsubsection{Sweeping Pruning Aggressiveness}

We conducted a sweep over pruning aggressiveness, varying the target sparse memory reduction from 10\% to 50\%. For each level of pruning, the model was retrained using the two pruning schedules. The percentile-based strategy served as a baseline pruning strategy, assigning higher pruning ratios to larger embedding tables. The FI-ITEP strategy dynamically adjusted pruning ratios according to feature importance scores.

Results indicate that both strategies achieved substantial memory savings, with total model memory reduction ranging from 5\% to 40\% depending on the pruning schedule. The FI-ITEP (both Shuffle and SVS) approach consistently outperformed the percentile baseline in terms of evaluation accuracy, demonstrating a more favorable trade-off between memory reduction and model performance.

\subsubsection{Repetitions on High Pruning Aggressiveness}

To assess the robustness of each pruning strategy, we performed 3–5 independent training runs at the highest pruning aggressiveness (targeting ~50\% sparse memory reduction). For each run, we recorded both training and evaluation normalized entropy (NE) metrics. The average of all FI-ITEP runs exhibited lower train NE compared to the percentile baseline, indicating greater stability under aggressive pruning. The average NE gap for the FI-based schedule was approximately 1.4\%, while the percentile baseline showed a regression of around 2.2\% compared to the baseline DLRM with no ITEP applied.

\subsubsection{Feature Importance Algorithm Comparison}

We further compared two feature importance algorithms—shuffle and Shapley value sampling—within the FI-ITEP framework on Recommendation Model 1. While both methods yielded comparable training NE, the shuffle-based approach demonstrated superior evaluation performance and lower variance across repetitions.

Tuning the number of batches for Shapley value computation presents a more intricate challenge compared to shuffle feature importance. While shuffle importance relies on repeated permutations to establish statistical robustness, the core principle remains relatively straightforward: more permutations generally yield more stable estimates. In contrast, Shapley value estimation (SVS) involves sampling from an exponentially growing space of feature coalitions. Determining an optimal batch size for this sampling requires a nuanced understanding of the trade-off between computational cost and the accuracy of marginal contribution estimates across diverse coalition structures. Inadequate batching can lead to biased or highly variable Shapley values, potentially misrepresenting feature contributions, a complexity not typically encountered to the same degree with shuffle importance. It's likely that this approach needs to run with far more batches to produce strong feature importance scores for the large-scale recommendation models explored in this work.

As a result, subsequent experiments with other recommendation models utilized the shuffle-based FI scores for pruning ratio configuration.

\subsection{Recommendation Model 2}

Recommendation Model 2 is a large DLRM that is over 1 TB in size with two separate recommendation tasks. We implemented aggressive sparse embedding pruning (up to 45\% sparse HBM reduction) which only moderately impacted model performance. On average, Percentile-ITEP yields 28\% HBM savings on this model, and FI-ITEP observed a total memory reduction of around 32\%. These results are summarized in Table \ref{tab:model_summary}.

\subsubsection{Repetitions on High Pruning Aggressiveness}
We perform the same 3–5 independent training runs at the highest pruning aggressiveness for this model. We evaluate the average train and evaluation performance over 3 repetitions of percentile ITEP and FI-ITEP on both the primary and secondary recommendation tasks of Recommendation Model 2.

These results highlight both the significantly improved train and eval NE accuracy of FI-ITEP over the percentile-based ITEP on this model. We discuss this promising direction in more detail in Section \ref{sec:overall_results}.

\subsection{Recommendation Model 3}

Recommendation Model 3 is a small sized DLRM just under 100 GB in size with a single recommendation task. We implemented aggressive sparse embedding pruning (up to 45\% sparse HBM reduction) which only moderately impacted model performance. On average, Percentile-ITEP yields 32.3\% HBM savings on this model, and FI-ITEP observed a total memory reduction of around 34.98\%. These results are summarized in Table \ref{tab:model_summary}.

\subsubsection{Repetitions on High Pruning Aggressiveness}

Results from three training repetitions with this small DLRM show that FI-ITEP outperformed the percentile baseline in terms of evaluation NE. While both strategies achieved substantial memory savings, both FI-ITEP and percentile pruning led to some NE regression compared to the unpruned baseline, indicating a loss in predictive accuracy under aggressive pruning on this smaller model. However, FI-ITEP still demonstrated reduced NE regression compared to the percentile approach. Across repetitions, FI-ITEP's eval NE regression (about 0.017\%) was smaller than the percentile baseline's (about 0.025\%).

\subsection{Overall Results}
\label{sec:overall_results}

For Recommendation Models 1 and 2, we observe that FI-ITEP performs the best of all tested pruning strategies, achieving both greater memory savings and improved evaluation NE compared to the percentile baseline. For Recommendation Model 3, FI-ITEP still outperforms the percentile baseline, though both approaches show some NE regression on this smaller model.

As model size surpasses some threshold, the model tends to learn embeddings through a greater diversity of features, many of which contribute minimally to overall model performance. The shuffle feature importance metrics on Recommendation Models 1 and 2 are more accurate in identifying the important features in these larger models, enabling more precise pruning decisions. This targeted reduction of less impactful embeddings not only streamlines the model, improving computational efficiency, but also acts as a form of regularization weighing more on features and embeddings with higher contribution to model quality, helping to mitigate overfitting and explaining why FI-ITEP performs best on these two model sizes.

On smaller models like Recommendation Model 3, the benefits of FI-ITEP are still present but less pronounced, as the model has fewer redundant embeddings to prune without impacting performance. We believe numeric contribution from feature importance is a criterion for characterizing embedding tables' learning capabilities, which assists in estimating and deciding pruning ratios for each table.

\section{Feature Importance Stability Over Time}
\label{sec:fi_stability}

We analyze how FI scores drift across retraining periods using Shapley-value FI sampling
(Table~\ref{tab:fi_drift}). Over a one-month interval, FI is highly stable (Spearman correlation of
the top-100 features between 0.85 and 0.92, with fewer than 12\% of top-100 positions changing).
Over two months, drift increases (correlations of 0.67--0.81, and 20--34\% of top-100 features
changing), most notably for the largest model (Model 2). In practice, large-scale recommender systems are
redeployed far more frequently than monthly to capture newer user signals, so we do not expect FI
drift to be problematic during deployment, and the GPU savings from pruning substantially outweigh
the periodic cost of recomputing FI.

\begin{table}[h]
\centering
\resizebox{\textwidth}{!}{%
\begin{tabular}{lcccccc}
\toprule
& \multicolumn{2}{c}{\textbf{Model 1}} & \multicolumn{2}{c}{\textbf{Model 2}} & \multicolumn{2}{c}{\textbf{Model 3}} \\
\cmidrule(lr){2-3}\cmidrule(lr){4-5}\cmidrule(lr){6-7}
\textbf{Metric} & 1 mo & 2 mo & 1 mo & 2 mo & 1 mo & 2 mo \\
\midrule
Top-100 FI score correlation (Spearman) & 0.90 & 0.78 & 0.85 & 0.67 & 0.92 & 0.81 \\
\% change in top-100 features                     & 9.5\% & 21.2\% & 11.8\% & 34.1\% & 8.9\% & 20.5\% \\
\bottomrule
\end{tabular}%
}
\vspace{2mm}
\caption{\textit{FI score drift (Shapley sampling) across one- and two-month retraining intervals.}}
\label{tab:fi_drift}
\end{table}

\end{document}